# OriWheelBot: An origami-wheeled robot

Jie Liu[1*], Zufeng Pang[2], Zhiyong Li[2], Guilin Wen[1*], Zhoucheng Su[3*],
Junfeng He[2], Kaiyue Liu[1], Dezheng Jiang[1], Zenan Li[1], Shouyan Chen[2],
Yang Tian[1], Yi Min Xie[4], Zhenpei Wang[3], Zhuangjian Liu[3]

[1]*Hebei Innovation Center for Equipment Lightweight Design and Manufacturing, School of Mechanical Engineering, Yanshan University, Hebei 066004, China*
[2]*School of Mechanical and Electric Engineering, Guangzhou University, Guangzhou 510006, China*
[3]*Engineering Mechanics Department, Institute of High Performance Computing, A\*STAR Research Entities, 1 Fusionopolis Way, 138632, Sinchannelore*
[4]*Centre for Innovative Structures and Materials, School of Engineering, RMIT University, Melbourne 3001, Australia*

*Correspondence should be addressed to
J.L. (email: jliu@ysu.edu.cn)
G.L.W. (email: glwen@ysu.edu.cn) OR
Z.C.S. (email: suzc@ihpc.a-star.edu.sg)

Origami-inspired robots with multiple advantages, such as being lightweight, requiring less assembly, and exhibiting exceptional deformability, have received substantial and sustained attention. However, the existing origami-inspired robots are usually of limited functionalities and developing feature-rich robots is very challenging. Here, we report an origami-wheeled robot (OriWheelBot) with variable width and outstanding sand walking versatility. The OriWheelBot's ability to adjust wheel width over obstacles is achieved by origami wheels made of Miura origami. An improved version, called iOriWheelBot, is also developed to automatically judge the width of the obstacles. Three actions, namely direct pass, variable width pass, and direct return, will be carried out depending on the width of the channel between the obstacles. We have identified two motion mechanisms, i.e., sand-digging and sand-pushing, with the latter being more conducive to walking on the sand. We have systematically examined numerous sand walking characteristics, including carrying loads, climbing a slope, walking on a slope, and navigating sand pits, small rocks, and sand traps. The OriWheelBot can change its width by 40%, has a loading-carrying ratio of 66.7% on flat sand and can climb a 17-degree sand incline. The OriWheelBot can be useful for planetary subsurface exploration and disaster area rescue.

**INTRODUCTION**

Terrestrial mobile robots equipped with different types of wheels to cope with complex environments have shown great application value in many fields [1,2]. Three common and effective forms are circular, tracked, and legged wheels [3-5]. The wheel rims of round-wheeled terrestrial mobile robots are continuous so that they can move on flat ground in a stable and high-speed manner, yet they cannot adapt to some uneven and complex environments, and their ability to cross obstacles is also limited. Tracked terrestrial mobile robots can adapt well to soft and wet terrain, but due to their complex structure, it is challenging to realize precise control, while the moving speed and efficiency could be much higher. Legged terrestrial mobile robots can cope with complex environments, and their obstacle-crossing ability is better than tracked ones. However, due to the discontinuity of the legs, it is prone to cause bumps in the moving process. The three aforementioned terrestrial mobile robots share a common characteristic, i.e., there is only a single mode, and the wheels cannot adapt to environmental changes, significantly hindering their engineering applications.

Multi-motion modes can be achieved by combing two or three of the above three types of wheels into terrestrial mobile robots [6-8]. For example, a wheel-legged robot primarily uses wheels in normal conditions, while switches to the legged mode when it encounters obstacles [6]. Terrestrial mobile robots equipped with hybrid circular-track-leg wheels can handle even more complex environments [7]. However, these robots suffer from certain drawbacks, including excessive components and intricate assembly procedures, leading to large, unwieldy robots with reduced reliability. There is a pressing need for these robots to advance toward higher performance and reduced weight to meet the diverse demands of various fields [9,10].

Incorporating origami techniques into robotics may offer desirable features like lightweight construction, simplified assembly, and adaptability, thus meeting the demands of various challenges [11-18]. Origami techniques empower robots with complex three-dimensional morphology changes and achieve rich functionalities through folding and unfolding processes [19,20]. Self-folding origami robots can be realized by integrating shape memory composites at the creases, with applications ranging from building self-assembling satellites in space to manufacturing centimeter-scale robots [21]. Origami techniques have also enabled smaller-scale robots, exemplified by the 90 mg flapping wing robotic insect [22]. Reconfigurable modular design concepts enhance mobility and maneuverability [23]. Nevertheless, a challenge arises as these robots, while capable of transitioning from flat configurations to complex three-dimensional structures for their intended functions, often possess monolithic motion models, limiting their adaptability to complex external environments. To address this, Lee et al. [24] proposed a variable-diameter wheeled robot inspired by a 'magic ball' origami, allowing the robot to adjust its overall height when encountering obstacles. Yet, its width and height are still coupled, sometimes unsuitable and challenging to control [24]. Inspired by the umbrella structure and origami techniques, Banerjee et al. developed a variable-diameter wheeled robot employing cable actuation, enabling it to traverse obstacles up to 5 cm in height at larger volumes [25]. Untethered-driven origami robots hold tremendous potential for engineering applications, accomplishing tasks like untethered walking, swimming, carrying objects, climbing slopes, crawling, and digging [18, 26-30]. Notably, the ability to navigate sandy terrain is crucial for robotic planetary explorations, such as missions to the Moon [31-34]. Existing planetary exploration robots rely on complex mechanisms and control systems to adapt to varying walking environments [35-37]. Consequently, there is a pressing need for engineering robots that are untethered, lightweight, require less assembly, possess adaptability, and offer multifunctionality.

To address these challenges, we introduce the OriWheelBot, a robot equipped with origami wheels that can change their width, and it possesses remarkable abilities to navigate sandy terrains. The origami wheels, inspired by the Miura origami, enable the variable width feature. We have developed analytical models that allow us to predict the width and radius of the origami wheels by utilizing the hypothetical condition of rigid folding. We have created two versions of the robot, the OriWheelBot, and an improved iteration called the iOriWheelBot. The latter boasts an automatic variable width function based on ultrasonic ranging. We demonstrate the untethered mobility and multifunctionality of these robots, showcasing their variable width feature, capability to traverse soft sand, transport objects, ascend slopes, walk on inclines, and navigate through sand pits, small rocks, and sand traps.

**RUSULTS**

**Construction of the origami-wheeled robot (OriWheelBot)**

The variable width origami wheel is inspired by the classical Miura origami. The Miura origami is constructed by periodically arranging its unit cell in the $x$ and $y$ directions. **Fig.1a** presents the four typical folding states of the Miura origami, where the folding angle $\theta$ gradually increases from left to right. The 2D geometrical pattern of one unit cell of Miura origami is determined by three independent parameters, i.e. a length $a$, a width $b$, and an acute angle $\beta$ (left panel of **Fig.1b**). One more parameter, i.e. the folding angle $\theta$, is needed to identify its spatial topology (right panel of **Fig.1b**). The 2D crease pattern of the origami wheel unit cell is the evolution from that of the Miura origami as shown in left panel of **Fig.1c**; it has three valley creases and one mountain crease. The parameter dominates the geometry of the 2D crease pattern of the unit cell becomes $l_t$, $l_u$, $b$, and $\beta$. $l_t$ and $l_u$ indicate the distance from the vertex, $T$, to the upper margin and the lower margin, respectively. $\beta$ is the acute angle formed by the right valley crease and the $x$-axis and $b$ the half width of the unit cell. Let the length of the unit cell be $a$, one gets $a = l_t + l_u$. One additional parameter, i.e. the folding angle, $\theta_w$, is required to determine the topology of the unit cell (right panel of **Fig.1c**). The four facets are marked as $S_1$, $S_2$, $S_3$, and $S_4$, respectively. To ensure the integrity of the origami wheel and facilitate installation of the hub later, two additional plates (marked as $S_5$ and $S_6$) are added to the unit cell. Periodically arraying the unit cell in the $x$ and $y$ directions forms the 2D crease pattern of the origami wheel; for instance, **Fig. 1d** shows a 2 × 8 arrangement. We perform the sequential folding at the predefined creases and connect the first unit cell to the last one into a closed circle, obtaining the 3D geometrical model of the origami wheel, as shown in **Fig.1e** ($S_{6\text{-all}}$ means all the $S_6$ in the origami wheel). Take the origami wheel as the basic unit, we assemble the OriWheelBot (**Fig.1f**). The OriWheelBot mainly includes six parts, i.e. four identical origami wheels (marked as **1**), the robot frame (marked as **4**), the variable-width driving device (marked as **2**), the mobile drive module (marked as **3**), the control module, and the battery module. Among them, the origami wheel is the most vital part of achieving the functionality of the OriWheelBot. The variable-width driving device drives the origami wheels to deform. The mobile drive module realizes the back-and-forth movement of the OriWheelBot. The control module is used as the controller for the two sorts of movements. The battery module is the powder supply for the OriWheelBot. The role of the robot frame is to carry the other five parts and other possible weights. One prototype of the OriWheelBot is fabricated and assembled (**Fig.1g** and Supplementary Note S1), and the fabrication process of the core component, origami wheel, is present in Supplementary Video S1. In addition, we have designed an improved version, labelled as iOriWheelBot. Relative to the current version, an automatic variable width function based on ultrasonic ranging has been added. The only difference in geometry between OriWheelBot and iOriWheelBot is the visual test module.

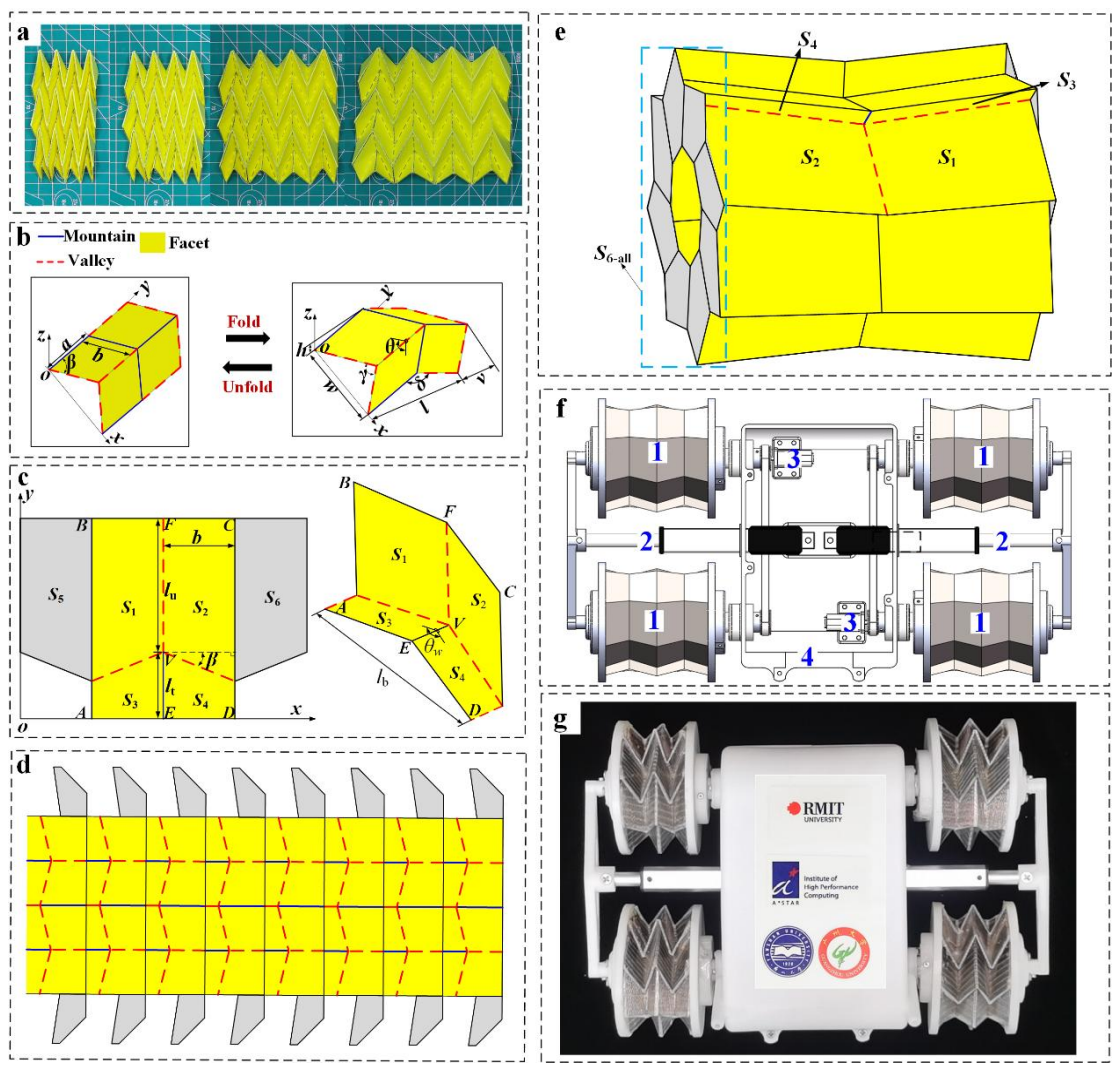

**FIG.1.** Construction of the OriWheelBot. (**a**) Four representative folding states of the classical Miura origami, in which the folding angle gradually increases from left to right. (**b**) The geometry of one unit cell of Miura origami (left panel: the 2D geometrical pattern; right panel: the 3D geometrical topology). (**c**) The geometry of one unit cell of the origami wheel (left panel: the 2D geometrical pattern; right panel: the 3D geometrical topology). (**d**) The 2D crease pattern of the origami wheel (two and eight

unit cells in the *x* and *y* directions, respectively, i.e., *n*=2, *m*=8). (**e**) The 3D geometrical model of the origami wheel. (**f**) Geometry of the OriWheelBot. (**g**) The assembled prototype of the OriWheelBot.

**Variable-width and diameter modelling**

The variable width capability is the essential functional requirement of the OriWheelBot, which is attributed to the foldability of the origami wheel. Thus, it is critically essential to model the variable-width capacity of the origami wheel. We are more concerned about two indicators, i.e. the width and the diameter of the origami wheel. The maximum and minimum of the origami wheel widths are vital to determine the obstacle avoidance ability of the OriWheelBot. By spatial geometric analysis, we derive the width, $l_b$, of the origami wheel, i.e., $l_b = 2mb\sin(\theta_1/2)$ (**Figs. 2a-2c**), where *m*, *b*, and $\theta_1$ are the number of the unit cells in the *x*-direction, the side length of parallelograms in the unit cell, and the folding angle, respectively. The minimum width $l_{b\min}$, and the maximum width $l_{b\max}$ of the origami wheel are then obtained. **Figs. 2a-2c** depict the spatial parameters, the schematic diagram of partial folding, and the cross-section schematic of the origami wheel, respectively. **Fig. 2a** gives $\varphi + \beta = \pi / 2$. In **Fig. 2b**, the blue dotted circle is the outer circle of the positive *n*-sided shape formed by the *n* origami wheel unit cells. $\gamma$ is the angle of the center of the circle contributed by a single origami wheel unit with $\gamma = 2\pi / n$. The diagonal valley crease in the unit cell can be extended to form a positive *n*-prismatic cone with an angle $\varphi$ between the prongs and the bottom edge. One can obtain the following geometric relations:

$$\cos\frac{\theta_1}{2} = \frac{(l_u - l_t)/2 \tan(\gamma/2)}{((l_u - l_t)\tan\varphi)/2} = \frac{1}{\tan(\gamma/2)\tan\varphi} \tag{1}$$

$$\theta_1 = 2\arccos\left(\frac{1}{\tan(\gamma/2)\tan\varphi}\right) \tag{2}$$

$$n > \frac{\pi}{\arctan((l_u - l_t)/2)} \tag{3}$$

We divide the radius into two sections, i.e., the radius of the solid, $r_1$, and the radius of the hollow, $r_2$, (**Fig. 2d**). *O* is the center of the origami wheel, *E* is a vertex on the periphery of the origami wheel, and *N* is the point where the *OE* line intersects the inner periphery of the origami wheel. Thus, we have $\overline{EN} = r_1$ and $\overline{ON} = r_2$. We derive the formulations of $r_1$ and $r_2$ via geometric analysis (**Fig. 2d** and Supplementary Note S2). The radius of the origami wheel is $r_d = r_1 + r_2$ and $\overline{OE} = r_d$.

$$r_d = \frac{b(\lambda_2 \sin(2\beta) + \cos(2\beta))\sqrt{x^2\lambda_2^2 + 1}}{\lambda_2 x \sin(2\beta) + \cos(2\beta)} + \frac{l_u - l_t}{2}\sqrt{(1-y)^2 + \left(\frac{1}{\tan\beta}\right)^2} \tag{4}$$

where $l_u = \overline{MG}$, $l_t = \overline{FG}$, $b$ is the half width of the unit cell, $\beta$ is the acute angle formed by the right valley crease and the $x$-axis (Fig. 1c), $\lambda_2 = l_t/b$, $x = l_1/l_t$, $y = 2l_2/(l_u - l_t)$, $l_1 = \overline{HG}$.

We conduct physical experiments to validate the effectiveness of the analytical solution for the origami wheel width and radius, each with five groups. Since the maximum value of the origami wheel width is our biggest concern, we compare the analytical value of this index with the physical experiment test value. **Fig. 2e** shows the origami wheel widths obtained from the five groups of experimental tests are 28.2, 21.0, 22.6, 35.4, and 33.3 mm, respectively, and the corresponding analytical solutions are 27.865, 20.403, 22.064, 34.831, and 32.863, respectively, finding that the relative errors are 1.2%, 2.9%, 2.4%, 1.6%, and 1.3% (The specific parameters of the origami wheels are shown in Table S2 of Supplementary Note S2). **Fig. 2f** depicts the test results for the origami wheel diameter (twice the radius) are 53.6, 53.4, 53.7, 51.1, and 55.7 mm, respectively, while the analytical predictions calculated by equation (4) are 52.915, 52.915, 52.915, 50.715, and 55.245 mm, showing a relative error of 1.3%, 0.9%, 1.5%, 0.8%, and 0.8%, respectively (The specific parameters of the origami wheels are summarized in Table S3 of Supplementary Note S2). Considering the no-thickness and rigid folding assumption in the derivation of the analytical solutions and the errors brought in during fabricating the origami wheel prototypes, these physical experiments well validate the accuracy of the analytical models of origami wheel diameter and width.

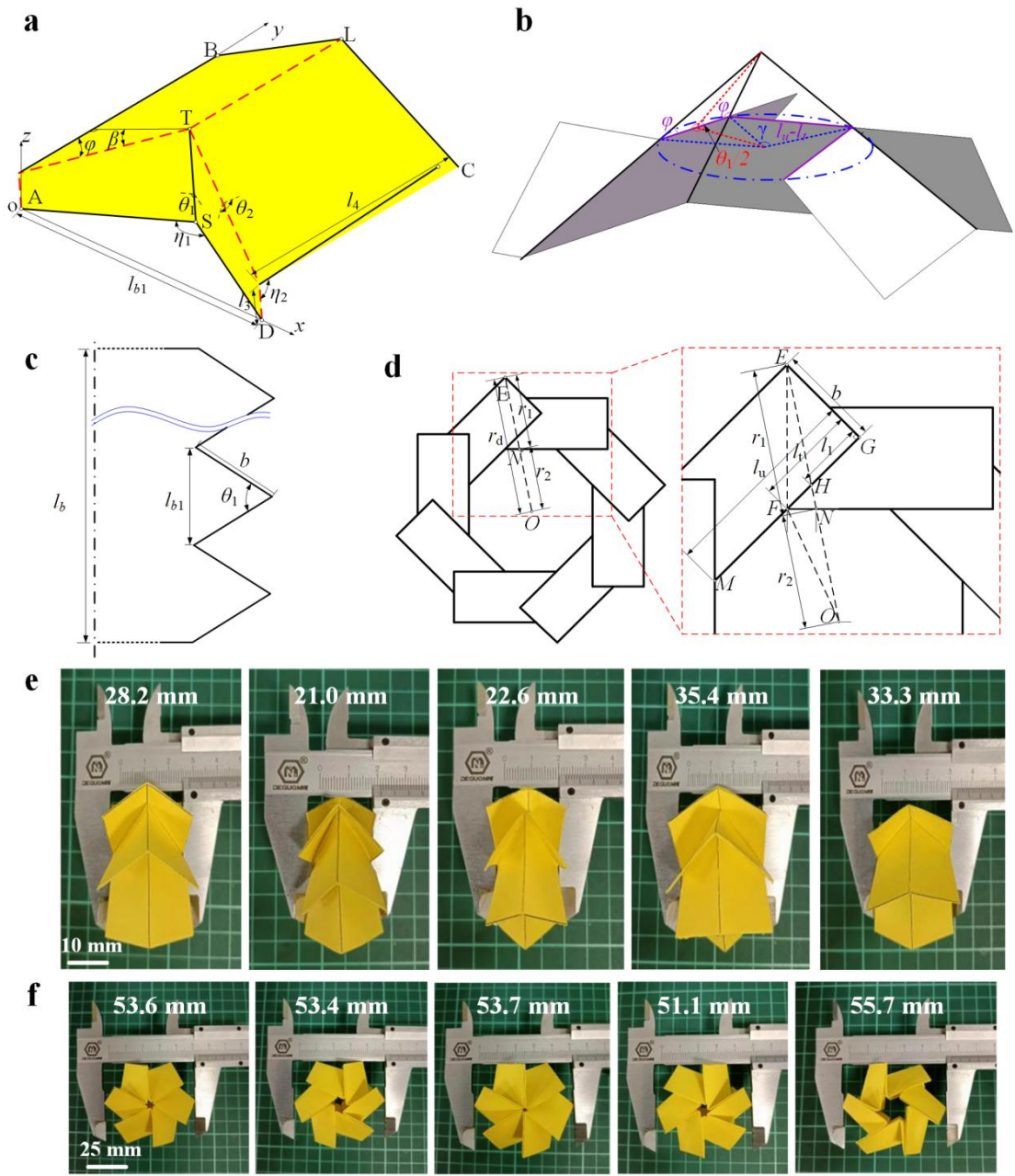

**FIG2.** Variable-width and diameter modelling of the origami wheel. (**a**) Spatial parameters of the origami wheel, which will be used to derive the width of the origami wheel. (**b**) Schematic diagram of partial folding of the origami wheel. (**c**) Origami wheel cross-section schematic. (**d**) The side view of an origami wheel (right panel: partial enlarged view), which will be employed to derive the radius of the origami wheel. (**e**) Five groups of physical tests are conducted to measure the width of the origami wheel to verify the analytical solution of the origami wheel width. (f) Five groups of physical tests are conducted to measure the radius of the origami wheel to validate the analytical solution of the origami wheel radius.

**Performance tests**

(1) **Variable width capacity.** We test this performance on a rubber pad with a thickness of 2 mm to maintain the unity of the test environment (**Fig. 3a**). When the OriWheelBot encounters a channel smaller than its width (**Initial state**), it folds the origami wheels to make its width smaller (**Origami wheel folding**), and then passes through the channel (**Through the channel**). After passing through the channel, the origami wheels are unfolded to their original state (**Back to the initial state**) because the OriWheelBot's walking is more stable in the case of the wider origami wheels (see **Fig. 3a** and Supplementary Video S2). The minimum channel that the OriWheelBot can pass is 247 mm, correspondingly, the minimum width of the origami wheel is 22 mm. The maximum width of the origami wheel is 72 mm when the OriWheelBot is in the initial state, and the overall width of the OriWheelBot is 347 mm. Thus, the width change ratio of the origami wheel can be up to 3.3 times, that is, the width of the wheel can be changed arbitrarily within the range of 1 to 3.3 times. The width change ratio of the OriWheelBot can reach 1.4 times, i.e., the robot can pass through a channel of 247 mm width when the width during stable operation is 347 mm. Supplementary Video S2 also presents the case when the OriWheelBot walks on the sand. We can further increase the ratio of the maximum width to the minimum width of the origami wheel by reprogramming the crease pattern of the origami wheel, thereby increasing the ratio of the maximum width to the minimum width of the OriWheelBot. **Figs. 3b-3d** present the variable width capacity of the iOriWheelBot (Supplementary Video S3), including three cases. Case 1: when the iOriWheelBot detects that the width of the channel is larger than the width of the iOriWheelBot, the iOriWheelBot does not need to change its width and passes directly through the channel (**Fig. 3b**). Case 2: when the iOriWheelBot detects that the width of the channel is less than the minimum width of the robot, the iOriWheelBot returns directly (**Fig. 3c**). Case 3: when the iOriWheelBot detects that the width of the channel is in the variable width range of the robot, the robot first performs origami wheel folding, then passes through the channel, and finally returns to the initial state (**Fig. 3d**). The components and the control process of the iOriWheelBot, as well as the control programme are presented in Supplementary Note S3.

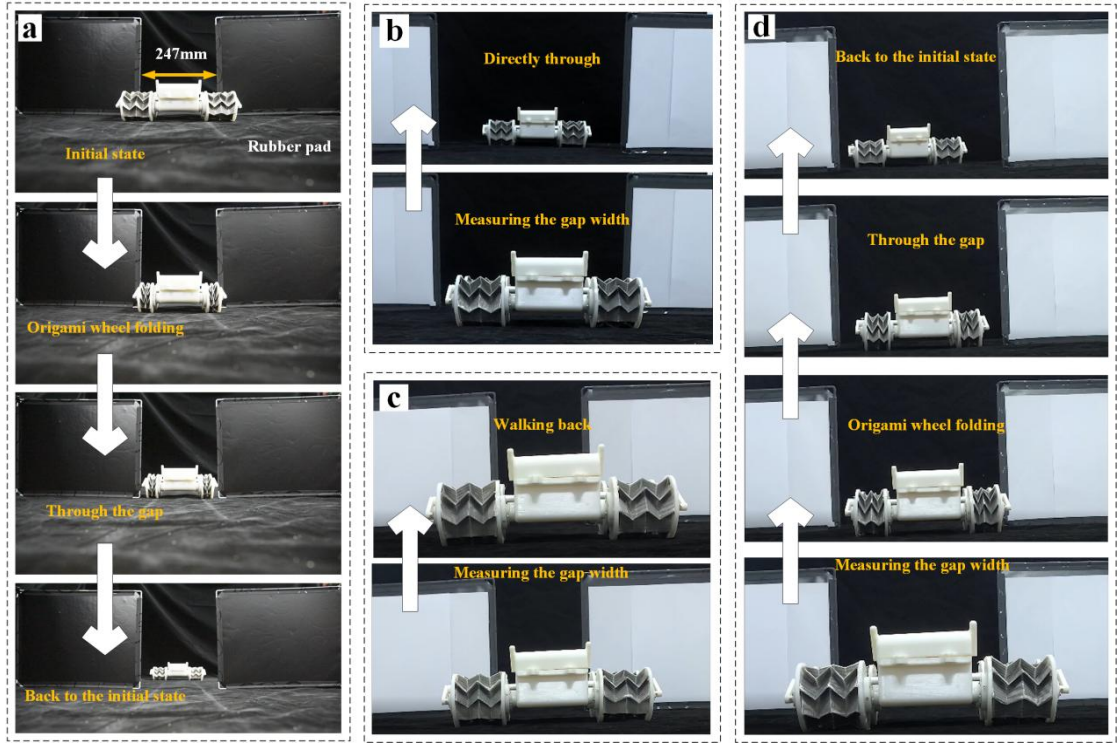

**FIG3.** Variable width capacity tests of the OriWheelBot and its improved version iOriWheelBot. (**a**) The variable width capacity of the OriWheelBot; when the OriWheelBot encounters a channel smaller than its width, it goes through four steps, i.e., initial state, origami wheel folding, through the channel, and back to the initial state. The variable width capacity of the iOriWheelBot: (**b**) When the iOriWheelBot detects that the width of the channel is larger than the width of the iOriWheelBot, the iOriWheelBot does not need to change its width and passes directly through the channel. (**c**) When the iOriWheelBot finds that the width of the channel is less than the minimum width of the robot, the iOriWheelBot returns directly. (**d**) When the iOriWheelBot detects that the width of the channel is in the variable width range of the robot, the robot first performs origami wheel folding, then passes through the channel, and finally returns to the initial state.

(2) **Soft sand walking ability.** We first investigate the ability to pass sand subsidence under the identical wheel turning and different contact areas with the sand. We find two motion mechanisms, i.e. sand-digging and sand-pushing mechanisms, when the origami wheels rotate clockwise and counterclockwise, respectively (**Fig. 4a**). When the origami wheel rotates clockwise in the soft sandy environment, the $S_1$, $S_2$, $S_3$, and $S_4$ of the unit cell form a closed sand dug as a groove for main contact with the sand. The digging trough regularly produces continuous sand-digging during its rotation. When the origami wheel rotates counterclockwise, only the closed $V$-shaped groove formed by $S_1$ and $S_2$ is used as the groove for main contact with the sand. Since $S_1$ and $S_2$ block $S_3$ and $S_4$, $S_3$ and $S_4$ are not in direct contact with sand. In this case, the *V*-shaped groove regularly produces continuous sand-pushing movement during the rotation process. Moreover, the ways of generating driving force under the two mechanisms are also different. In the sand-digging mechanism, first, the origami wheels first enter the sand with the connecting mountain fold line between each wheel folding unit; until the sand completely fills the closed sand-digging trough and contacts $S_1$, $S_2$, $S_3$, and $S_4$, and then through the rotation of the wheels, $S_3$ and $S_4$ generate a direct thrust on the sand. At the same time, the sand produces resistance to the origami wheels, that is, 'line contact→surface contact→generating driving force'. While in the sand-pushing mechanism, the *V*-shaped groove directly contacts the sand and generates an interaction force, which is a process of 'surface contact→generating driving force'. We observe the track of the origami wheel on the sand is *W*-shaped and *M*-shaped, respectively (**Fig. 4b** and Supplementary Video S4). In the sand-digging mechanism, we find a small amount of sand accumulates in the sand chute after leaving the sand layer, which may throw sand into the air along the direction of rotation, thereby causing tiny sand particles to enter the robot, and in turn affects the movement the OriWheelBot. Nevertheless, this issue is not found in the sand-pushing mechanism. This is owing to the sand dug trough faces away from the sand during the process of pushing sand, and the sand in the dug trough moves along the $S_1$ and $S_2$ poured vertically into the sand to avoid the impact of sand on the OriWheelBot parts. Thus, from this point of view, we are more inclined to the sand-pushing mechanism.

We study the capacity of the OriWheelBot with different wheel widths to sink through the soft sand using the sand-pushing mechanism. The initial position of all working conditions is that the OriWheelBot sinks 30 mm, which is the distance from the chassis to the lowest point of the origami wheel. The OriWheelBot with three typical wheel widths are selected, i.e. 22 mm (minimum width), 38 mm, and 72 mm (maximum width). We test two indicators, i.e., the time required to pass through the sand subsidence (marked as $t_{pt}$) and the position after passing through sand subsidence at a specific time, e.g., 8 s (marked as $d_{pt}$). We find $t_{pt}$ is roughly equal to 6, 4, and 1 s and $d_{pt}$ approximately equals 60, 130, and 330 mm for these three cases (**Fig. 4c**). We reveal that the width of the origami wheel turns larger, the time for the OriWheelBot to pass through the sand subsidence becomes shorter, and the passing ability through the sand subsidence gets stronger. This can be attributed to the fact that a larger origami wheel diameter corresponds to a larger contact area between the origami wheel and the sand, which increases friction and thus enhances the sand-walking ability of the OriWheelBot.

We develop finite element (FE) models to support the above observations (Supplementary Note S4). Since the OriWheelBot's ability to walk in sand is primarily a function of wheel-sand interaction, we focus on qualitatively investigating the interaction of a single origami wheel with sand. **Fig. 4d** shows the element void fraction of the different models at the end of the FE simulations, indicating the sand profiles. As can be seen, the sand profile is perturbed less and less as the wheel width increases. For $l_b$ = 22 mm, the sand is violently perturbed as the origami wheel sinks into the sand significantly and the sand is pushed backwards and upwards forming piles behind the wheel. For $l_b$ = 38 mm, the origami wheel sinks into the sand after it travels by around 150 mm, the pile is formed as the wheel starts to sink quickly. For $l_b$ = 72 mm, the wheel hub sinks into the sand forming the shallow grooves. The interaction between the origami sheet and sand does not create too much change to the sand profile. Figs. **4e-4h** show the vertical displacement versus the travel distance curves for different origami wheel widths, $l_b$ = 22, 38, and 72 mm, under the identical friction coefficient $\mu$ = 0.2. It is clearly seen that the wheel digs into the sand very significantly for $\omega$ = 50 rad/s. For $l_b$ = 22 mm, the origami wheel sinks into the sand quickly until it touches the bottom of the Eulerian domain. This is mainly because the wheel is too narrow, and the sand is unable to support the wheel with $\omega$ = 50 rad/s. The sand is pulled up by the origami sheets. With relatively smaller angular velocity, i.e., $\omega$ = 6.28 rad/s, the sand can support the origami wheel longer. However, the origami wheel still sinks significantly into the sand and the OriWheelBot cannot walk with such angular velocity. For $l_b$ = 38 mm and $l_b$ = 72 mm, the sand can support the origami wheel even if $\omega$ = 50 rad/s. However, the reaction from the sand is too large that the origami wheel tends to climb out of the sand. Especially for $l_b$ = 72 mm, the wheel is bounced off the sand because of the transient dynamic impact on the sand. It is shown that steady stage movement can be achieved with $\omega$ = 6.28 rad/s (Fig. **4h**). It is also observed that the maximum depth the origami wheel can reach into the sand decreases as the origami wheel width increases if the velocity and friction coefficient are the same. This agrees with experimental results. Other FE simulations results with different friction coefficients are shown in Supplementary Note S5.

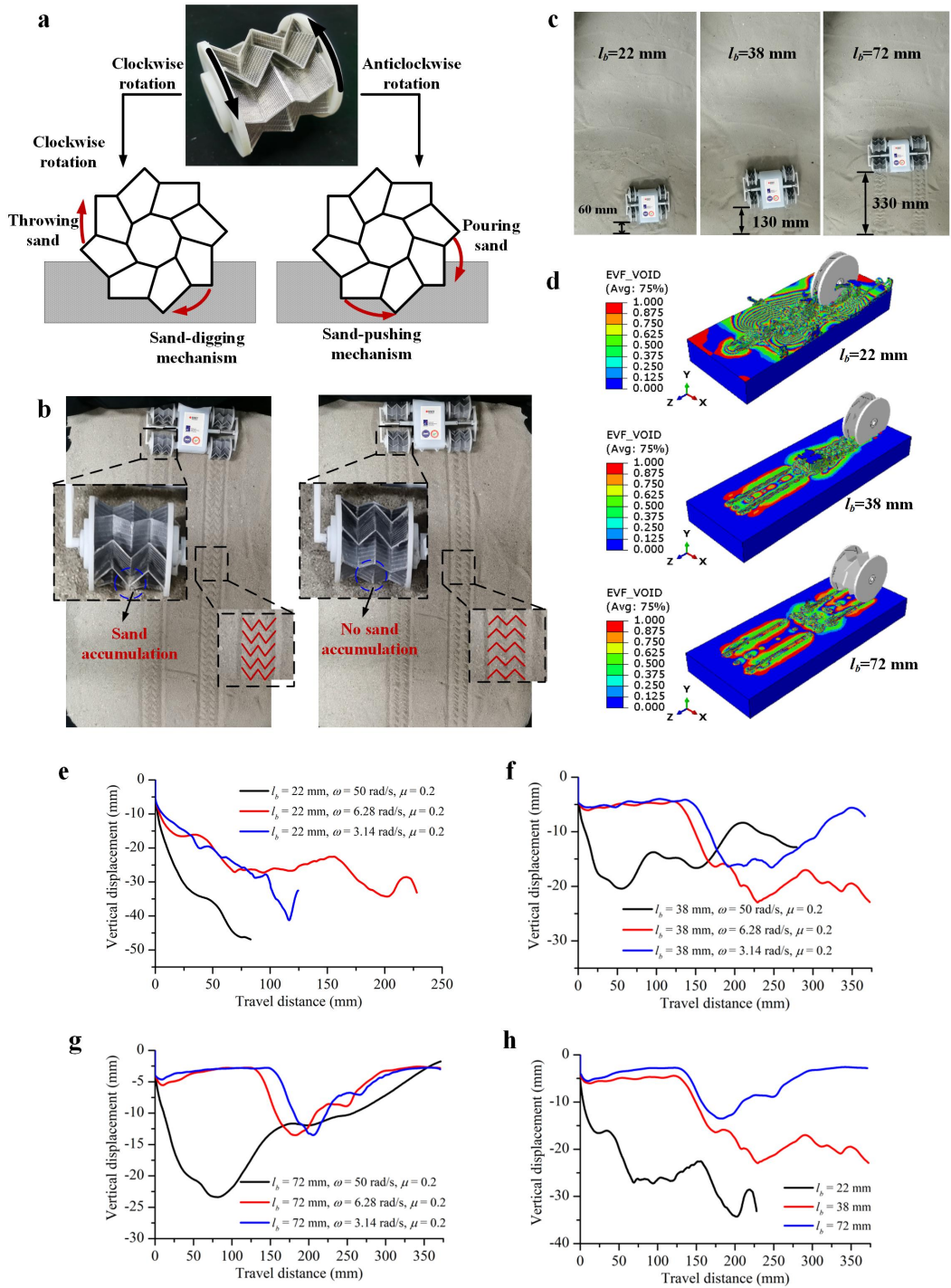

**FIG4.** Soft sand walking ability of the OriWheelBot. (**a**) Two motion mechanisms, i.e., sand-digging and sand-pushing mechanisms, when the OriWheelBot walks in the soft sand. (**b**) The walking state when the OriWheelBot walks in the soft sand (left panel: sand-digging mechanism and right panel: sand-pushing mechanism). (**c**) The capacity of the OriWheelBot with different wheel widths to sink through the soft sand by using the sand-pushing mechanism (from left to right $l_b$ equals 22, 38, and

72mm). (**d**) Element void fraction of the different models at the end of the simulations. (**e**)-(**h**) The vertical displacement versus the travel distance curves for different origami wheel widths, $l_b$ = 22, 38, and 72 mm, under the identical friction coefficient $\mu$ = 0.2; $\omega$ = 6.28 rad/s for (**h**).

**(3) Other superior performance in sand.** We further test other superior performance in the sand of the OriWheelBot, including carrying blocks, climbing a slope, walking on a slope, passing through a sand pit, small rocks, and sand traps. We test the ratio of the load-carrying capacity ($M_{LBC}$) of the OriWheelBot to its weight ($M_{OriWheelBot}$ = 1348.5 g) for the cases when $l_b$ = 22, 38, 55, and 72 mm, respectively. The loading-carrying ratio ($\rho = M_{LBC}/M_{OriWheelBot}*100\%$) is measured by adding weights to the OriWheelBot until it cannot move and taking this weight as the ultimate load. We identify that $M_{LBC}$ roughly equals 250 g, 450 g, 700 g, and 900 g, achieving a loading-carrying ratio of 18.5%, 33.4%, 51.9 %, and 66.7% (**Fig. 5a** and Supplementary Video S5). We conclude that the larger the origami wheel width, the higher the loading-carrying ratio. This is because enlarging the origami wheel width will increase the contact area of the origami wheel and the sand, further enhancing the load-carrying capacity. We observe that compared with the sand-digging mechanism, the sand-pushing mechanism has unique advantages in sand slope climbing (with a slope of 17 degree; see the inserted panel of **Fig. 5b**). The OriWheelBot does not obtain enough driving force to pass through the sand slope through the sand-digging mechanism. Instead, the sand-digging mechanism will form the sandpit, causing the OriWheelBot to sink at the sand slope. In this situation, the robot frame will come into contact with the sand; as a result, the OriWheelBot will get stuck, further preventing it from climbing the sand slope. On the contrary, no sandpit is formed under the origami wheel in the sand-pushing mechanism. Although the OriWheelBot is slightly sinking in this situation, the robot frame and the sand slope still maintain a certain channel, negligible influencing the OriWheelBot's climbing ability in the sand. Therefore, identical to the soft sand walking ability test, we use the sand-pushing mechanism to investigate the OriWheelBot's climbing ability in the sand. Through experience and exploration, we design a sand slope with a slope of about 17 degree. We investigate the OriWheelBot's climbing ability for the sand slope for the aforementioned cases. We find that: when $l_b$ = 22 mm, the OriWheelBot's front two wheels fail to pass the top of the sand slope. This is due to the contact area between the origami wheel and the sand being very small, unable to provide sufficient driving force, and there is a phenomenon that the frame is stuck in the sand; for the cases when $l_b$ = 38 and 55 mm, the contact area between the origami wheel and the sand becomes larger, the OriWheelBot's front two wheels pass the top of the sand slope. However, owing to the insufficient contact area of the wheels, the two rear origami wheels sink into the sand slope, causing the OriWheelBot still cannot climb the sand slope smoothly; for $l_b$ = 72 mm, the contact area is at the largest, and the OriWheelBot has the ability to pass through the sand slope completely (**Fig. 5b** and Supplementary Video S6). **Fig. 5c** shows the trajectories of the origami wheel on a sand slope with a 17-degree inclination by using numerical simulations (Supplementary Video S7). It can be found that when the width is 38 mm, the origami wheel sinks deeper and deeper into the sand as the OriWheelBot walks (about 25 mm at 100 mm, as shown in **Fig. 5d**) until it can no longer walk.When the width is 72 mm, the OriWheelBot can vary from sinking 5 to 15 mm cycles (**Fig. 5d**), confirming that it can complete the climbing of that slope well. These simulation results can qualitatively corroborate the experimental observations in **Fig. 5b**.

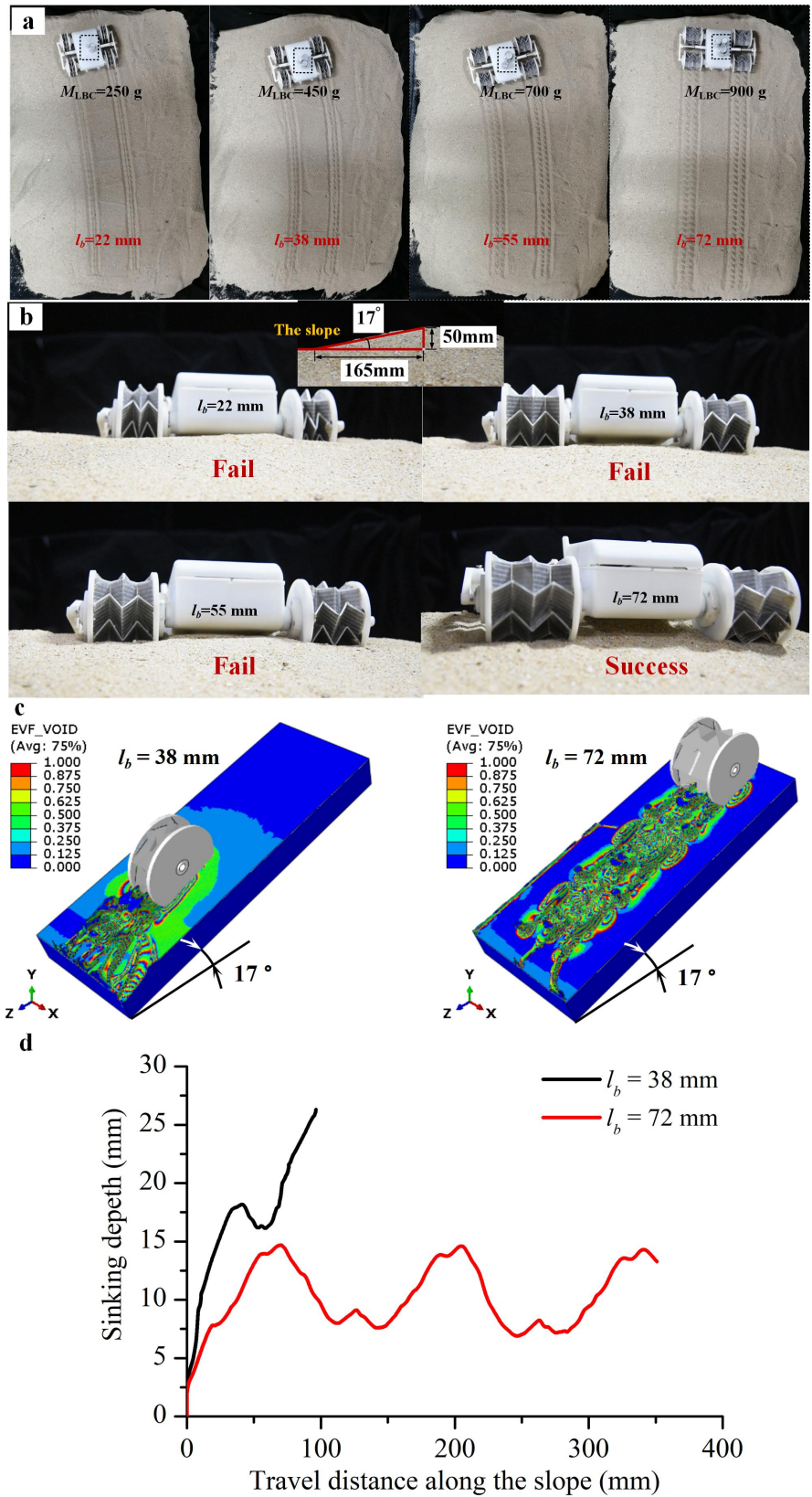

**FIG5.** The carrying blocks and climbing capacities on the sand. (**a**) The load-carrying capacity on soft sand with different wheel widths (using weights to character $M_{LBC}$). (**b**) Climbing ability on the soft sand; when $l_b$ equals 22, 38, and 52 mm, the OriWheelBot fails to climb the 17 degree slope, and the OriWheelBot with the wheel width of 72 mm can complete the mission. (**c**) Trajectories of the origami wheel on a 17 degree slope when $l_b$ = 38 mm (Left panel) and $l_b$ = 72 mm (Right panel). (**d**) Sinking of the origami wheel into the sand.

    **Figs. 6a-6d** demonstrate the power of the OriWheelBot to operate in potential environments in areas such as planetary subsurface exploration. **Fig. 6a** shows that the OriWheelBot with a wide origami wheel can successfully walk on a sand slope (Supplementary Video S8). We present two states, i.e., states A and B, to demonstrate the two different moments of the OriWheelBot. **Fig. 6b** reveals the ability of the OriWheelBot to pass through a sand pit. The OriWheelBot first accesses the slope of the sand pit, then reaches the inside, and finally escapes from the sand pit (Supplementary Video S9). We have labelled the walking paths with blue arrows. We place some small stones on the sand and let the OriWheelBot pass over them, finding that the OriWheelBot can pass over them very well (**Fig. 6c** and Supplementary Video S10). Finally, we elaborately arrange some small traps in the sand to examine the OriWheelBot's ability to pass. It can be found that the OriWheelBot can pass over small sand traps smoothly (**Fig. 6d** and Supplementary Video S11).

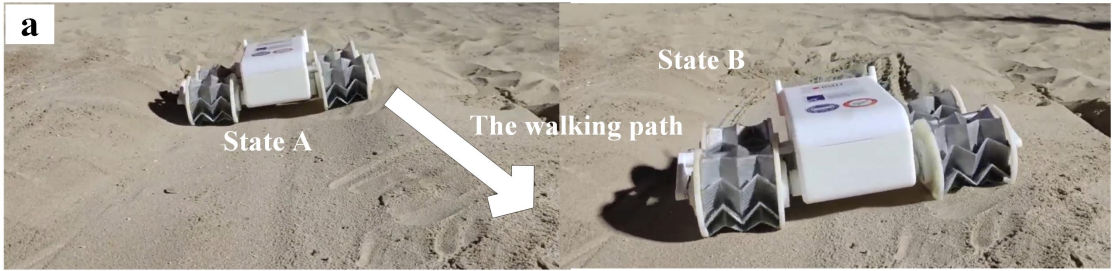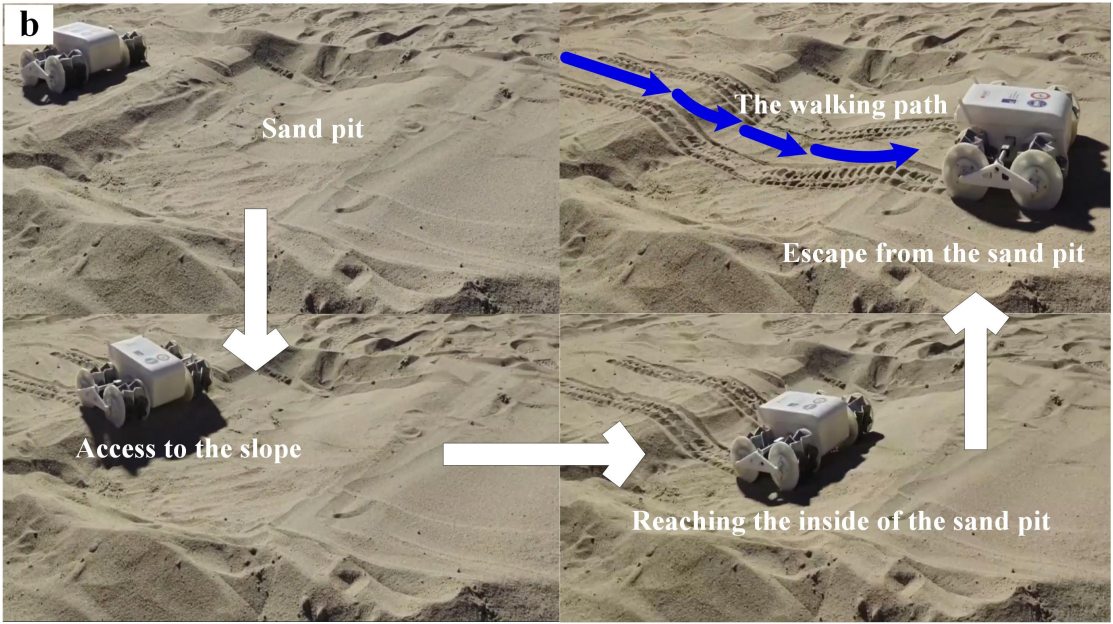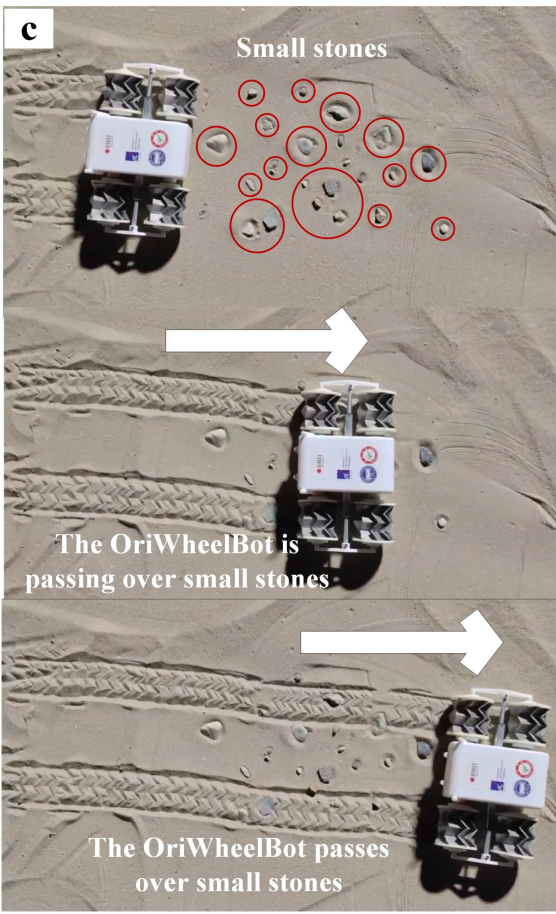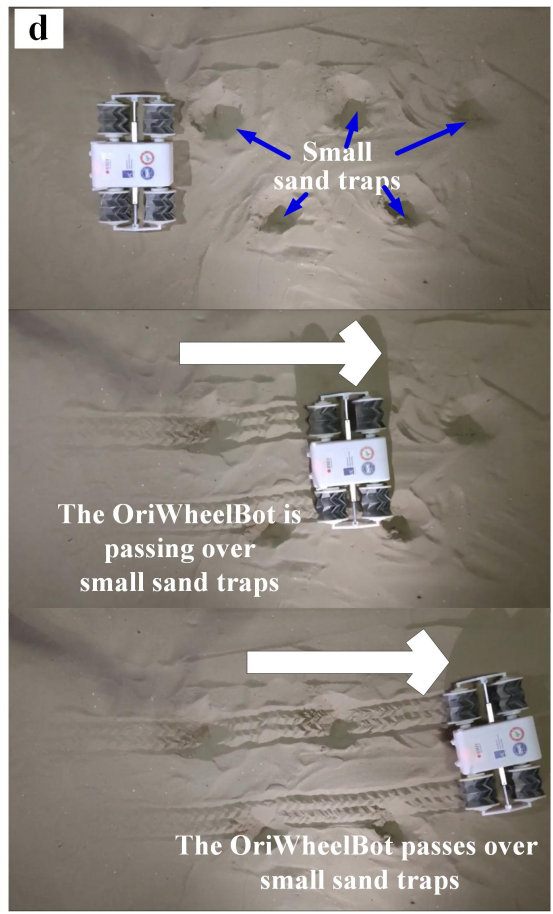

**FIG6.** Other superior performance in the sand. (**a**) Walking on a sand slope: two states are demonstrated, i.e., state A and B. (**b**) The ability of the OriWheelBot to pass through a sand pit: it first accesses the slope of the sand pit, then reaches the inside, and finally escapes from the sand pit. (**c**) Passing over small rocks. (d) The OriWheelBot can pass over small sand traps.

**CONCLUSIONS**

In this study we introduce the OriWheelBot, a robotic platform incorporating origami-inspired variable-width wheels derived from the Miura origami. We undertake a comprehensive analysis of the origami wheel's width and radius through both analytical and experimental methods, with a particular focus on enhancing obstacle-crossing capabilities. The OriWheelBot offers several notable advantages, including lightweight construction, untethered operation, and simplified assembly processes. Additionally, we present an upgraded version known as the iOriWheelBot, equipped with automatic distance measurement and obstacle avoidance functionalities. Through a series of experiments and numerical models, we highlight the OriWheelBot's remarkable performance in sand-related tasks. These include navigating soft sand terrains, carrying loads, scaling slopes, walking on inclined surfaces, and successfully traversing challenging landscapes such as sand pits, small rocks, and sand traps. The OriWheelBot, as an untethered origami robot, holds significant potential for applications in planetary explorations.

Although the current OriWheelBot prototype features only two unit cells in the width direction ($n = 2$), it is important to mention that our design approach is readily scalable to accommodate a greater number of unit cells. **Fig. 7a** illustrates the conceptual design of the OriWheelBot in four scenarios with varying $n$ values (2, 4, 8, and 16), demonstrating its adaptability to different configurations. Further work is to develop the third generation of the OriWheelBot, which will be equipped with high-performance sensors, power supplies, smart material actuators, etc., and consider radiation and other extreme environments [31,38-42]. For example, **Figs. 7a and 7b** demonstrate the conceptual diagram of OriWheelBot applications in planetary exploration and disaster area rescue, respectively.

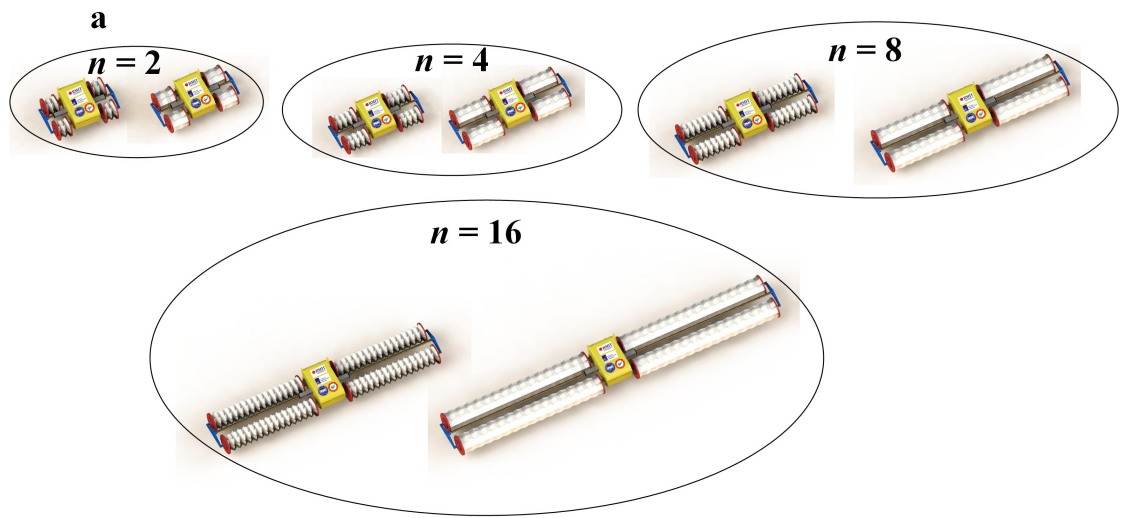

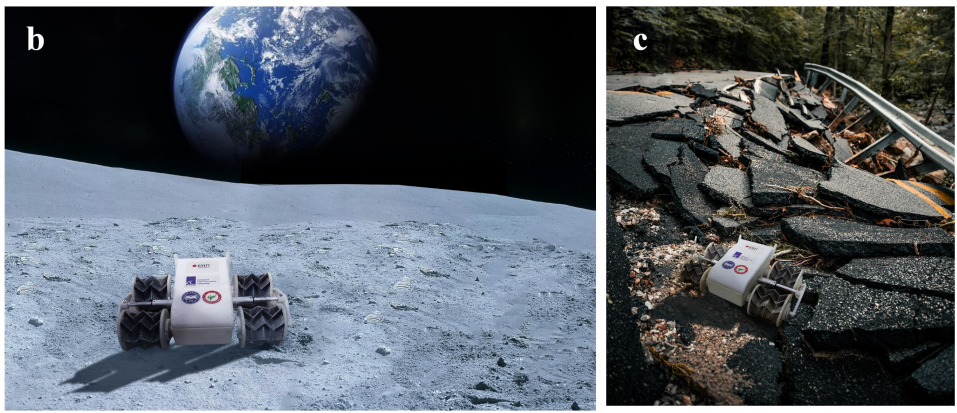

**FIG7.** Conceptual diagram of the OriWheelBot to demonstrate its (**a**) scalability; applications in (**b**) planet exploration and (**c**) disaster area rescue.

## Competing Interests

The authors declare no competing interests.

## Acknowledgements

This research was financially supported by the National Natural Science Foundation of China (Nos.11902085, 12172095).

## Supplementary material

Supplementary Note S1-S5
Supplementary Figure S1-S8
Supplementary Table S1-S4
Supplementary Video S1-S11